# Development of REGAI: Rubric Enabled Generative Artificial Intelligence

Zach Johnson
Department of Computer Science
North Dakota State University
1320 Albrecht Blvd., Room 258
Fargo, ND 58108

P: +1 (701) 231-8562
F: +1 (701) 231-8255
E: zachary.e.johnson@ndsu.edu

Jeremy Straub
Department of Computer Science
North Dakota State University
1320 Albrecht Blvd., Room 258
Fargo, ND 58108

P: +1 (701) 231-8196
F: +1 (701) 231-8255
E: jeremy.straub@ndsu.edu

**Abstract**

This paper presents and evaluates a new retrieval augmented generation (RAG) and large language model (LLM)-based artificial intelligence (AI) technique: rubric enabled generative artificial intelligence (REGAI). REGAI uses rubrics, which can be created manually or automatically by the system, to enhance the performance of LLMs for evaluation purposes. REGAI improves on the performance of both classical LLMs and RAG-based LLM techniques. This paper describes REGAI, presents data regarding its performance and discusses several possible application areas for the technology.

**<u>Keywords:</u>** rubric, generative artificial intelligence, AI, GAI, large language model, LLM, retrieval augmented generation

# 1. Introduction

The advent of large language models (LLMs), like ChatGPT 3.5 and 4.0, has sparked the growth of artificial intelligence (AI) applications for both personal and business uses [1]. These models excel in various language-related tasks including summarization, essay writing, and creative content production [2]. However, their adoption in critical business applications has been hindered by concerns over hallucinations - instances where LLMs generate responses that are disconnected from the prompt, factually incorrect, or nonsensical [3][4].

As the use of generative AI (GAI) in business settings grows, regulatory bodies are focusing on its implications and potential risks. Recent developments in AI regulation aim to ensure responsible and ethical use of AI systems in commercial applications. For instance, the European Union's AI Act includes stringent rules for high-risk AI applications, such as employment decisions and educational assessments [5]. In the United States, agencies like the Federal Trade Commission are scrutinizing the use of AI in business to prevent deceptive or unfair uses [6]. These regulatory frameworks emphasize the need for transparency, accountability, and robust evaluation mechanisms in AI systems – particularly those that impact critical decision-making processes in business environments.

LLMs often struggle with tasks requiring human-like reasoning and planning capabilities [7]. While their outputs may appear superficially plausible, they frequently lack the logical coherence and depth of human-generated content. This limitation has led to limited use of LLMs in non-chatbot business applications [8][9][10], despite the significant economic potential in automating tasks that require human reasoning and language skills.

To address these challenges, rubric enabled generative artificial intelligence (REGAI), a technology for evaluating unstructured text data, was developed. REGAI aims to closely align LLM outputs with human reasoning and interpretation capabilities, making it suitable for a wide range of applications including academic grading, job candidate evaluation, and professional-client interaction assessment. The REGAI system is comprised of several key components: an (optionally) LLM-generated, human expert-reviewed rubric that defines detailed criteria for text evaluation; a scoring LLM that generates initial scores based on the rubric and examples of other scores; a critiquing LLM, trained on human critiques, which reviews and provides feedback on the initial scoring; and an iterative draft-critique cycle that refines the evaluation until it meets defined quality criteria.

REGAI incorporates several features to enhance its performance and alignment with human judgment. These include retrieval-augmented generation (RAG), a curated knowledge base of expert evaluations [11], and a self-strengthening mechanism that continuously improves the system's performance through expert validation. Flexible tuning mechanisms are also incorporated, including few-shot and one-shot prompt engineering, to align the system's outputs with human-generated evaluations [1]. REGAI addresses critical challenges in automated text evaluation, such as maintaining consistency, reducing subjectivity, and scaling to large volumes of data [12]. By combining the strengths of LLMs with human expertise, REGAI offers a potential solution for high-volume, objective, and structured text evaluation across various domains.

This paper continues with Section 2, which provides background on rubrics, generative AI, RAG, and existing LLM-human alignment methods. Then, Section 3 presents a comprehensive overview of the REGAI system. Section 4 explores potential applications for the technology, while Section 5 outlines key requirements for its implementation. Section 6 evaluates the technology's performance. Finally, section 7 presents the conclusions drawn from this study and gives suggestions for areas of future work.

REGAI aims to bridge the gap between LLM capabilities and the complex reasoning required in real-world text evaluation tasks. It offers the potential for improving how businesses and institutions approach large-scale document assessment and decision-making processes.

## 2. Background

This section provides an overview of key concepts and technologies relevant to the development of advanced text evaluation systems. It focuses on rubrics, generative AI, retrieval augmented generation, and previous attempts at aligning LLMs with human judgment.

### 2.1. Rubrics

Rubrics are structured scoring guides used to evaluate performance in various domains, such as education and professional assessment. They typically consist of a set of criteria and descriptors for different levels of performance for each category. They also typically include a scoring system [13].

The use of rubrics has been shown to increase the reliability and validity of assessment processes [14]. They provide a standardized framework for evaluation, reducing subjectivity and enhancing consistency among raters [15]. In education, rubrics have been found to improve student learning by clarifying expectations and facilitating targeted feedback [16].

Recent research has explored the potential of automated rubric-based LLM assessment systems. They have been studied for evaluating workshop courses led by experts [17], essay feedback generation [18], and evaluating student computer programming assignments [19]. These systems aim to apply rubric criteria consistently to large volumes of work, addressing challenges of scale in assessment tasks [12]. However, the complexity of interpreting nuanced criteria and applying them to diverse texts remains a significant challenge for automated systems.

### 2.2. Generative AI

Generative artificial intelligence systems create new content, such as text, images, and other media. In natural language processing, generative AI has delivered advancements in the development of large language models such as generative pre-trained transformers (GPTs) [1], bidirectional encoder representations from transformers (BERT) [20], and their derivatives.

These models are trained with large amounts of text data and use sophisticated neural network architectures, primarily based on the transformer model [21]. Transformers revolutionized natural language processing with their self-attention mechanism. This mechanism allows the model to weigh the importance of the different parts of an input sequence when processing each element, enabling more effective handling of long-range dependencies in text [22]. Transformers are the foundation for most state-of-the-art language models due to their scalability and performance [23].

LLMs have demonstrated impressive capabilities in various areas such as text generation, summarization, translation, and question-answering [2]; however, they also present significant challenges. These include the potential for generating biased or inappropriate content [7], difficulties in maintaining factual accuracy [3], and the phenomenon of hallucination, where models generate plausible but incorrect information [4].

Additionally, regulatory issues have emerged as a critical concern. Governments and organizations worldwide are grappling with how to regulate AI systems to ensure safety, privacy, and ethical use. For instance, the European Union has categorized AI systems based on their risk and imposes stricter regulations on high-risk applications [5]. United States agencies are exploring regulations to address AI fairness, transparency, and accountability [6].

## 2.3. Retrieval Augmented Generation

RAG is a hybrid approach that combines the strengths of retrieval-based and generation-based natural language processing techniques. It enhances the performance of generative models by incorporating relevant information retrieved from a knowledge base during the generation process [11].

In a RAG system, when a query or prompt is received, the model retrieves relevant documents or passages from a large corpus. This information is used to condition the language model's output, allowing it to generate responses that are fluent and grounded in accurate and relevant information [24].

RAG has shown promising results in improving the accuracy and relevance of generated text, particularly in knowledge-intensive tasks such as open-domain question answering and fact verification [11]. It mitigates some of the limitations of generative models, such as hallucinations and a lack of up-to-date information. RAG models have demonstrated significant improvement over traditional language models. For instance, in testing using a natural questions dataset [11], RAG achieved a top-1 accuracy of 44.5%, compared to 32.6% for a T5-base model and 29.8% for BART-large. Similarly, with the WebQuestions dataset, RAG provided a 4 to 5 percentage point improvement in exact match accuracy over baseline models [11].

## 2.4. Previous Attempts at LLM Alignment with Human Scoring/Judging

Aligning the output of LLMs with human judgment is critical for the use of LLM-based AI systems for evaluation and decision-making tasks [25]. Several approaches have been explored to do this. One approach is fine-tuning on human-labeled data. To do this, models are trained on datasets that include human-generated scores or judgments. This approach has been used for automated essay scoring systems [26] and sentiment analysis tasks [27].

Prompt engineering is another technique. For this technique, researchers explored sophisticated prompting methods to guide LLMs towards producing outputs that better align with human judgment [28]. This includes few-shot learning approaches and carefully crafted instructions [1] [29].

Reinforcement learning from human feedback trains language models using rewards derived from human preferences [30]. It has been applied to tasks such as summarization and dialogue generation and shows improved output quality and alignment with human preferences [25]. Ensemble methods, which combine multiple models or incorporate rule-based systems alongside neural models, have also been used to improve alignment with human judgment [28] [31]. Additionally, explainable AI techniques have been employed to make LLM decisions more interpretable, including attention visualization and generating explanations alongside predictions [32][33].

Despite these efforts, achieving consistent alignment between LLM output and human judgment remains a significant challenge. This is particularly true in complex evaluation tasks that require nuanced understanding and reasoning [34].

## 2.5. Challenges with Automated Text Evaluation

Automated text evaluation presents several key challenges in automatically assessing written content [35]. Contextual understanding is a significant hurdle. Capturing the nuances of context, tone, and implicit meaning in text remains difficult for AI systems [31][36]. This is particularly problematic when evaluating complex written works where meaning often extends beyond literal interpretation [37].

Subjectivity in evaluation poses another significant challenge. Many evaluation tasks involve subjective elements that are challenging to quantify and automate [38]. This is especially pronounced with creative writing or opinion documents. Human evaluators will often disagree regarding these documents, making automated assessment difficult [39].

The diversity of writing styles also presents a complex challenge for automated systems [40]. Maintaining consistent evaluation criteria across various writing styles and structures is difficult [41]. Variability in language makes it difficult to create a one-size-fits-all evaluation system, as different styles may require different assessment approaches [42].

Ethical considerations, including issues of fairness, bias, and transparency in automated evaluation systems, are ongoing concerns [43][44]. These challenges are growing in importance as automated systems increasingly influence high-stakes decisions in education, employment, and other areas [45]. Ensuring that these systems do not create, perpetuate, or exacerbate existing biases is crucial for their ethical implementation [46].

The foregoing challenges demonstrate the need for innovative approaches that can leverage the strengths of both AI and human expertise in text evaluation tasks [47]. Addressing these issues is crucial for developing reliable and fair automated text evaluation systems that can be trusted for numerous applications [20].

## 3. System Overview

This section presents an overview of REGAI. It details each step of the process from rubric generation to final scoring, including the iterative critique cycle. Figure 1 provides a high-level overview of the processes presented in this section. A comprehensive diagram of the system is shown in Appendix A.

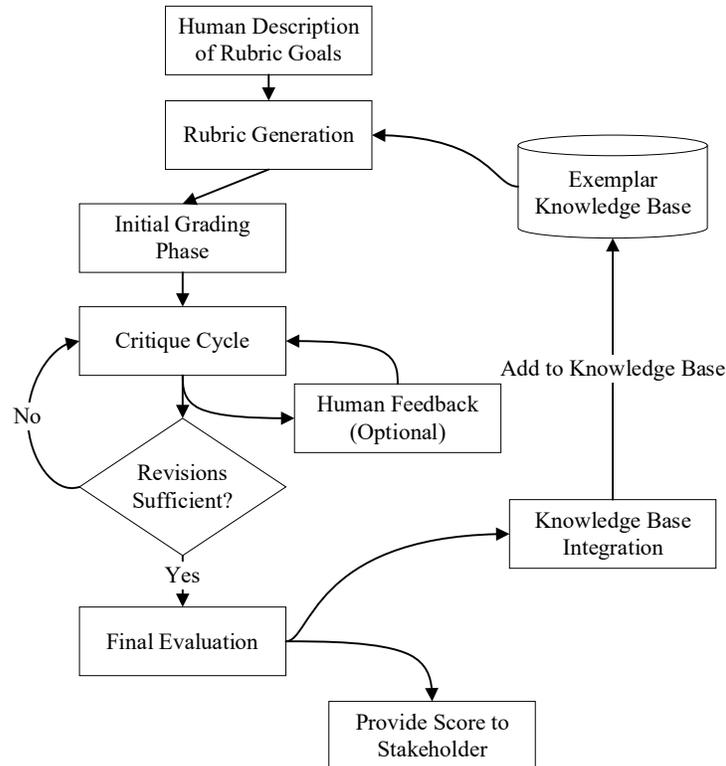

**Figure 1.** High-level overview of the REGAI pipeline.

## 3.1. Rubric Generation

The first step in the REGAI pipeline is the generation of a rubric that will be used to evaluate text data. This can be done manually or with assistance from the system.

If working with an expert, it is not necessary to include rubric auto-generation, if the expert is confident in their ability to provide scoring criteria for the task at hand. This is especially true for educational professionals, where rubric development will likely be a task they have done before.

However, for many experts, creating rubrics to frame the exact goals of a task may not be intuitive. Therefore, it may be necessary to provide a starter rubric that provides ideas for how to accurately frame the task in terms of a rubric.

If the expert does not need a starter rubric, they supply an initial rubric to the pipeline to be used for scoring. This may be a common situation once the expert is used to using the system or has a database of similar rubrics. Otherwise, if expert would like a starter rubric, the process proceeds as is described in this section. Figure 2 provides an overview of the rubric generation process. Each step is now discussed.

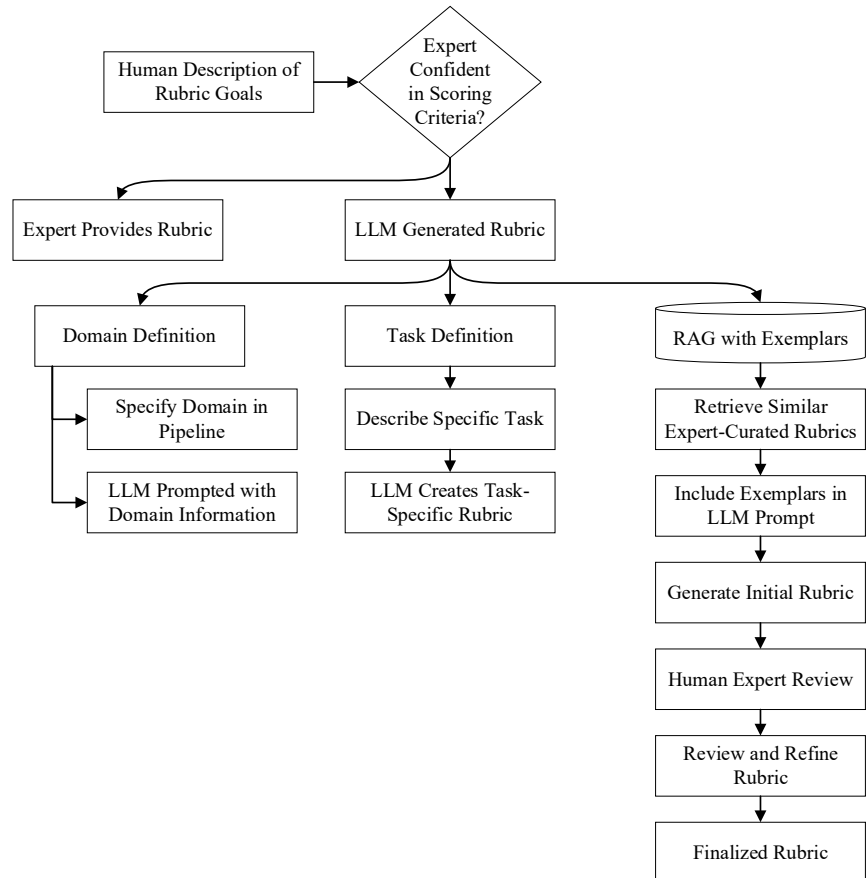

**Figure 2.** Process for the generation of an initial rubric.

*Domain Definition* – Prior to rubric generation, the domain of knowledge and task at hand should be stated as a prompt to the rubric-generating LLM (e.g., "you are an LLM assistant that creates rubrics for evaluating [description]"). This is typically included in the building of the pipeline, which is made solely for a specific task (e.g., grading English assignments or evaluating job candidates based on job descriptions). The rubric generation prompt given to the LLM will make it clear what the task is.

A potential alternative is a domain-agnostic system which asks the expert for a short description of the ranking task they would like completed, thereby defining the domain manually prior to rubric generation. This work focuses on optimizing domain-specific pipelines.

*Task Definition* - Once the domain is specified, the specifics of the task need to be provided. For instance, once the LLM knows it is supposed to make a rubric to grade English assignments

(domain), it needs a description of the assignment (specific task). Similarly, an LLM for making rubrics to evaluate job candidates must be given a job description on which to base the rubric.

The task description complexity varies by domain. In some areas, like email ranking, the criteria for scoring the incoming data consists of common knowledge which is already incorporated into the learned weights of the LLM. In these cases, it may not be necessary to provide a comprehensive task description. However, in other domains, like job matching, the nebulous task of creating a rubric to score job candidates necessitates that the task be defined by a job description. There is no way to specify, for instance, how to score a candidate's qualifications, skills, or education against the specific requirements listed of a job description, if it is not given.

*RAG Using Rubric Exemplars -* Finally, the prompt given to the LLM to initiate rubric creation should include exemplars. These are rubrics that have been previously created by experts for similar tasks. The document database is utilized for similarity comparisons and consists of key-value pairs of the form: {[task description prompt]: [expert created rubric]}. Once the task description prompt for rubric generation is made, it is indexed against all keys in the document store. The record with the highest cosine similarity is selected and this rubric is included as a one-shot example in the prompt.

Alternately, another selection measure could be used. For instance, the evaluation metric could be the product or geometric mean of a performance score (like average satisfaction rating, accuracy, or mean square error) and a similarity measure (like cosine similarity, L2 distance, or dot product similarity) such that performance and similarity are both weighed in the selection of an exemplar.

An expert rubric which was created for a similar purpose to the one being developed will be used as an example to the LLM in creating the new rubric. Ideally, this will lead to close adherence to the human-generated level of detail and clarity included in the new rubric.

The document database includes rubrics which have been verified to either have a close match with human judgements or which have been rated as exceptionally satisfactory. If a similarity threshold is not met, then the example rubric is not included to avoid confusing the rubric-generating LLM with a non-relevant example rubric. A consequence of this process, which ends in the addition of the rubric to the knowledge base (pending human review and empirical testing), is that, as more rubrics are created and tested, more rubrics will be in the document store. This increases the strength of the system as it is used more, making it a self-reinforcing system.

### 3.2. Human Expert Review

Once a rubric is generated, it is important that it be reviewed by a human expert. This step is crucial for aligning the process with human judgement. If the rubric has logical errors in it, any records scored using it will receive inaccurate evaluations.

The proposed system has functionality to allow rubrics to be revised, and records to be re-scored. This is especially useful in cases where an error is discovered in the rubric. In this process, a human can add columns or rows to the rubric, as they see fit, and assign new scores to the rows or weights to the columns. The weighing of each column is important as it allows the expert to assign value to certain attributes over others.

### 3.3. Initial Scoring Phase

The initial scoring phase is a critical step in the REGAI pipeline. During this phase, the system applies the human-reviewed rubric to evaluate the input text. This phase leverages the power of large language models and the accumulated knowledge from previous evaluations. The process is designed to be robust, accurate, and continuously improving. It sets the stage for the subsequent critique and refinement cycles. Figure 3 depicts this phase.

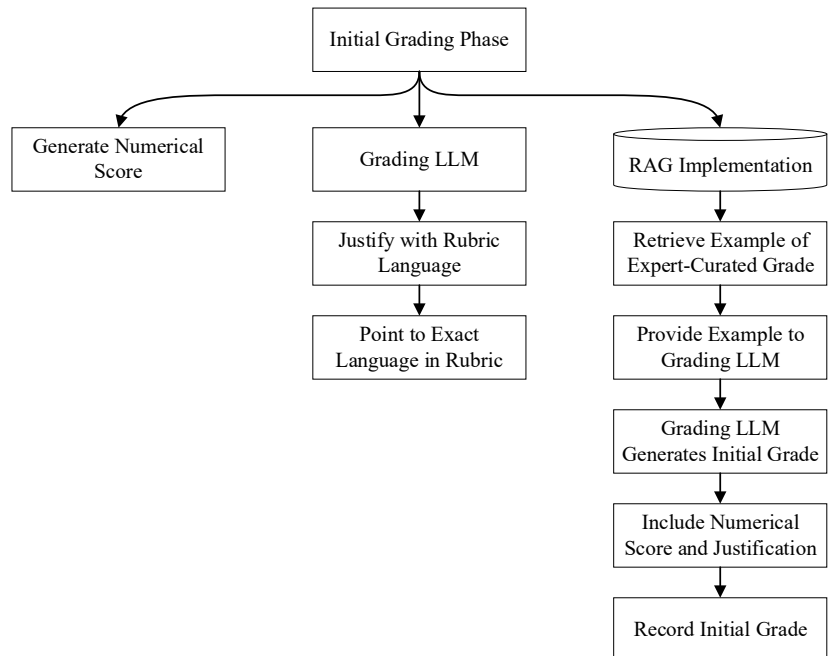

**Figure 3.** Initial Scoring Phase.

*The Scoring Agent* – In this phase, an initial score is given to a record. The score is produced by the scoring LLM, which should be a large model like GPT-4 turbo or Llama-3-70B. The score must include both a numerical component and a detailed justification in terms of each category in the rubric. The scoring LLM is instructed to point to exact langue in the rubric that corresponds to attributes of the text that led it to assign the score. This is not an easy task for an LLM to perform correctly, as it may require chain of thought and human-level reasoning in some areas (although the rubric minimizes the need for this). This is why the critiquing agent is necessary.

*RAG For Scoring Examples* – RAG is used, at this stage, to provide an example expert-curated score which was given to a similar record. The strength of this portion of the pipeline increases as the size and quality of the expert-informed knowledge base increases. Thus, it is self-reinforcing. The self-strengthening nature of this pipeline provides a competitive advantage for users that can build up large and/or high quality knowledge bases of expert rubrics and evaluations.

***Critique Cycle*** – The critique cycle is a crucial component of REGAI. It is designed to refine and improve the initial scoring output. This iterative process mimics the real-world interaction between an experienced evaluator and a novice scorer, ensuring that the final assessment adheres closely to the established rubric and maintains high standards of accuracy and fairness. The cycle involves a review-revise interaction between the scorer and critic, leveraging the strengths of large language models for both generative and analytical tasks.

***Critiquing Agent (Critic)*** – The critic reviews the output of the scoring module and explains to the scoring agent what it must fix. This is akin to a professor providing feedback to a student who is grading assignments for their course. Sometimes, the scorer does an excellent job of following the rubric and assigning the correct score to an assignment. In this case, the grade points to specific areas in the assignment that line up with specific scoring criteria in the rubric. In other cases, the scorer fails to consider certain details in its evaluation. These omissions need to be pointed out so corrections can be made. In other cases, the grader fails to include multiple details, which need to be corrected. This may require multiple rounds of correction to result in a fair and objective evaluation. Once the evaluation is deemed fair by the instructor, the score can be given to the student. This is the basic concept of the critique cycle.

***RAG for Criticism Examples*** – In the critique cycle, a RAG mechanism (like the one used in rubric generation) is implemented to find records most like the current one within the knowledge base. These are provided to the critiquing agent as exemplars to follow. Ideally, the knowledge base consists of expert-made critiques of initial LLM outputs. This knowledge base is created by asking experts to identify where LLM output did not align well with rubrics that they made.

Alternatively, a mechanism could be provided to display outputs to the user as they are generated and ask for a detailed description of the rubric's performance. In some commercial settings, however, users may not provide sufficient detail to ensure that scores conform to the rubric. In these cases, it can be useful to include multiple similar examples that are returned by the similarity search via a cosine similarity (or other) top-k algorithm. This provides the critic with exposure to a wider range of scores, which will hopefully align well with the current scoring task.

***Nature of Critique vs Generation Tasks*** – The sophistication and accuracy of the critic's critique is key. It is believed that the task of pointing out where an agent missed the mark, with respect to a rubric, is easier for an LLM agent than the task of following the scoring policy. Thus, the critic will give accurate and direct critiques that the scorer can act on to correct its evaluation. This component in the pipeline results in an alignment bottleneck. It forces the system back to following the human-specified rubric, if it has strayed from that task.

Critiquing is simpler for an LLM because it involves identifying deviations from a clear standard. The critiquing agent analyzes the scoring agent's output, detecting inconsistencies, and provides feedback. This leverages the LLM's pattern recognition capabilities, as the agent only needs to compare the scoring results against predefined criteria and point out discrepancies.

Identifying what is incorrect or missing is typically less cognitively demanding than the creative and interpretive tasks involved in scoring. The critic, equipped with the rubric and expert-created examples, pinpoints areas of improvement with a high degree of accuracy, ensuring that the scorer receives precise and actionable feedback.

This stands in contrast to the generation task assigned to the scoring agent that involves synthesizing information and applying abstract rules from the rubric to produce a coherent and contextually accurate score and justification. Scoring requires a higher level of comprehension and the ability to interpret and apply nuanced guidelines to diverse pieces of text. The scoring agent must not only understand the rubric's criteria but also translate these criteria into specific judgments about the text. This demands a deep understanding of the content and the ability to apply human-like reasoning. This makes the scoring task inherently more complex and challenging than critiquing, as it involves more cognitive load and has more potential for variability in interpretation. Once the scoring agent is given critiques of its output, it is a less creative and subjective task to simply apply those changes.

***Revision and Iteration*** – Using the critique, the scorer changes the rubric to correct the mistakes identified by the critic. This is a relatively simple process. The scorer is provided the critique as a user message and is instructed to return scores and justifications that correct the mistakes that were identified.

A new score set is produced by the scorer. This is sent to the critic for review and possibly additional feedback. The critic can choose to pass the scores or suggest further revisions. Along with the revised set of scores and justifications, RAG is used to provide the critic with examples of rubrics that pass a set of suggested revisions, and examples where further revisions are needed. The entire chain of revisions is included in the messages sent to the critic so it can see how the conversation evolved. Eventually, the scores should pass the critic's review. This will only happen when the scorer provides the requisite level of detail and reasoning in the score justification.

## 3.5. Use of and Addition to the Knowledge Base

Once the final scores have been created, the system must decide whether the text produced by the LLM should be added to the knowledge base to serve as an example for future scoring.

A key-value pair consisting of {[text to be evaluated]: [document]} format is used. Documents considered for knowledge base inclusion include the initial human-created or human-reviewed rubric, and the final score (and justifications) given to a document. A second knowledge base stores key-value pairs of the form {[draft score generated by scorer (with text under evaluation)]: [critiques given by critic]}. This knowledge base is used during the critique cycle to give the critic effective examples to reference.

The goal of this step is twofold. First, it seeks to add the highest performing pipeline documents to the knowledge base to increase the quality of the knowledge base. Second, it seeks to increase the volume of documents such that, over time, the similarity between text being analyzed and documents in the knowledge base is increased.

The extent to which a comprehensive knowledge base increases system accuracy is unknown. If the similarity of the text in the knowledge base to the text being evaluated is important, a large knowledge base will increase system accuracy.

Another possibility, however, is that the details of the text being evaluated are not as important as the critiques and level of detail that is included in the submitted scores and justifications. In this case, a small number of well-crafted examples in the knowledge base, which exemplify the necessary attributes of the pipeline but are not specific to all possible text, may be sufficient.

A third (unlikely) possibility is that a large knowledge base, which always provides examples from pieces of text like the one being scored, may cause the scorer to not generalize the scores and critiques from the example. The scorer may, instead, copy the scores and critiques to the new text, failing to evaluate the nuance of the unscored text. This is a key area of future study.

## 4. System Uses

This section discusses several prospective uses for REGAI. These application areas include academic grading, applicant application evaluation, performance evaluation, interaction evaluation, proposal evaluation, email screening, paper review and the review of court and interrogation transcripts. Each is now discussed.

### 4.1. Academic Assignment Grading

REGAI can automate the grading process for academic assignments, ensuring that the grader adheres closely to the instructor-created rubric. The system features a portal for instructors to submit assignment descriptions and rubrics. There is an option to create rubrics collaboratively, if they are not already available. Instructors can grade sample assignments to provide a baseline for the system and offer critiques of LLM-generated grades to refine the process. Due to FERPA considerations, the system may need to be implemented on-premises using a specially secured server.

This system can significantly improve grading objectivity, reduce the time spent on grading, and provide more specific and actionable feedback to students. An appeal process allows students to challenge grades, with the instructor making final decisions. While currently limited to writing assignments, this approach could potentially enhance academic assessment by combining AI efficiency with human oversight.

### 4.2. Human Resources: Applicant Tracking Systems

The accurate evaluation and ranking of job candidates, based on resumes, is a challenging task for hiring managers. This is especially true in sectors with high applicant volumes. Current methods often rely on keyword searches or parsed resume data, which can be gamed by applicants. This can advance those who game they system while also leading to qualified candidates being overlooked. REGAI addresses this by providing a fair and thorough evaluation of all applicants, based on a detailed rubric crafted by the hirer. The system allows hirers to review and critique scores, ensuring human oversight of the process.

REGAI mitigates the risk of unexplainable AI decisions in hiring by providing full justifications for all scores. This allows hirers to understand and, if necessary, override decisions. This addresses a key limitation of other applicant scoring methods.

### 4.3. Human Resources: Performance Evaluation

REGAI can assist human resources staff in evaluating employee performance. The ability to evaluate employee performance equally for all employees can facilitate cross-unit comparisons and reduce bias complaints.

To do this, REGAI is provided employee performance data in natural language. Many companies conduct weekly or daily stand-up meetings where employees say what they have done in the previous week or day, and what they plan to do the next. These could be recorded and used, in aggregate, for review.

The scoring process would follow the one described for applicant evaluation. A manager first defines a rubric to score employees' collection of language-described completed tasks. Examples are provided to serve as the initial knowledge base. Additional natural language performance data can be collected from other sources and assessed. Feedback can be provided to tune the system to deal with the subjectivity inherent in quantifying work.

### 4.4. Client Interaction Evaluation

REGAI can be used for evaluating interactions between professionals-in-training and simulated or real clients. This may be particularly valuable in fields such as medicine, law, therapy, and customer service, where interpersonal skills and domain knowledge are crucial. By developing rubrics that capture the nuances of effective communication, empathy, and technical competence, the system can provide consistent and objective feedback on trainee performance. This can decrease the cost of evaluation and allow more resources to be devoted to providing targeted training to those showing the greatest need.

For instance, in medical education, REGAI could be used to analyze recordings of student-patient interactions during clinical simulations. The rubric could assess history-taking skills, bedside manner, and diagnostic reasoning. This automated evaluation could supplement human instructor feedback, allowing for more frequent practice opportunities, faster feedback, and personalized learning experiences.

### 4.5. Solicited Proposal Evaluation Screening

Organizations that receive and evaluate proposals, such as grant-making agencies, can benefit from REGAI's capabilities for screening and making a preliminary ranking of submissions. The system can use rubrics that reflect the organization's priorities and evaluation criteria. Past successful proposals can be provided as exemplars. If novelty is required, RAG can be used to identify similar previously submitted proposals. Pair-wise comparisons can be performed by the pipeline.

By processing large volumes of proposals quickly and consistently, REGAI can help identify the most promising candidates for human review. This saves time and ensures that proposals receive a standardized initial assessment. The system can flag proposals that meet certain thresholds for immediate human attention, while providing feedback on others that may need revision before further consideration.

### 4.6. Email Filtering and Prioritization

Many email systems employ basic filtering algorithms. REGAI offers a more sophisticated approach to email management. By developing rubrics that capture the nuances of email importance, urgency, and relevance to the user's role and priorities, the system can provide an intelligent sorting mechanism.

REGAI can analyze incoming emails based on factors such as sender relationship, content relevance, action items, and time sensitivity. This allows for more granular categorization beyond simple spam detection, potentially creating dynamic folders or priority levels that adapt to the user's changing needs over time.

### 4.7. Technical Paper Review

The peer review process for academic and technical papers is time-consuming and subject to variability in reviewer assessments. REGAI can assist in this process by providing an initial evaluation of submitted papers, based on rubrics that reflect journal or conference standards. These rubrics might include criteria such as originality, methodology, clarity of presentation, and potential impact.

While not replacing human reviewers, the system could offer a standardized baseline assessment, highlight potential areas of concern, and even suggest relevant literature that the authors may have overlooked. Publications and conferences could allow authors to provide an initial revision in response to REGAI's feedback, improving the quality of papers reaching human reviewers. This could streamline the review process, reduce workload on human reviewers, and potentially improve the consistency of initial screenings and paper quality.

### 4.8. Court Case Transcript Analysis

The legal system generates vast amounts of court transcript text, which often requires careful analysis for purposes such as appeal preparation, legal research, and precedent identification. REGAI could be employed to analyze transcripts based on rubrics designed to identify key legal arguments, procedural issues, and potential grounds for appeal.

The system could flag important sections of testimony, highlight inconsistencies, or categorize arguments according to relevant areas of law. This would assist legal professionals in quickly navigating lengthy transcripts and identifying critical information, potentially improving the efficiency of legal processes and the quality of legal analysis.

## 4.9. Signals Intelligence and Interrogation Transcript Evaluation

Law enforcement and intelligence agencies often need to analyze large volumes of signals intelligence recordings and interrogation transcripts to extract valuable information and assess the credibility of statements. REGAI can potentially be used for this by crafting rubrics to identify indications of deception, consistency in narratives, and the presence of key information.

The system can help prioritize which data requires immediate human attention, identify patterns across multiple interrogations, and flag potential breaches of interrogation protocols.

## 5. Key Requirements for Use

The effective use of REGAI depends on several critical factors. This section outlines two key requirements: tracking model changes over time and incorporating explainability mechanisms.

## 5.1. Tracking Model Changes Over Time

As REGAI relies heavily on large language models (LLMs) and their outputs, it is crucial to implement a robust system for tracking changes in these models over time. This requirement stems from several factors inherent to the nature of LLMs and the REGAI system.

First, model updates must be tracked. LLMs like GPT-4 are frequently updated by their developers. These updates can introduce changes in model behavior, potentially altering the output quality or consistency of the REGAI pipeline. A tracking system allows for the identification of any significant shifts in performance that coincide with model updates. Alternatively, the pipeline could use static models and perform thorough testing prior to any migration to new updates.

Second, the knowledge base will evolve over time. REGAI is a self-reinforcing system and adds high-quality outputs to its knowledge base. This necessitates careful monitoring of how the evolving dataset impacts system performance. Tracking over time can identify trends in output quality and consistency as the knowledge base grows.

Third, domain drift my occur. In some applications, the nature of the text being evaluated may change over time due to shifts in language use, industry standards, or societal norms. Tracking model performance longitudinally helps detect when the system's effectiveness declines due to these factors. This phenomenon typically occurs over a much longer time scale than the other items discussed.

Fourth, calibration of the critique cycle is required over time. The effectiveness of the critique cycle, a key component of REGAI, may change over time as the scoring and critiquing agents evolve through exposure to more examples. Tracking these changes can inform adjustments to the critique process to maintain its alignment with human expert judgment.

To achieve these goals, several measures should be put in place.

First, regular benchmarking must be performed. To do this, a set of standardized test cases that can be run at regular intervals to assess system performance consistency can be established.

Second, version control must be maintained. To achieve this, detailed records of all components of the REGAI pipeline, including the specific versions of LLMs used, the state of the knowledge base, and any changes to prompts or rubrics should be maintained.

Third, there must be continuous performance metric logging. This is achieved by logging key performance indicators, such as agreement rates with human experts, processing times, and the number of critique cycles required for convergence.

Fourth, anomaly detection should be implemented. To do this, a system to flag sudden or unexpected changes in output patterns or quality metrics can be created.

Firth, human-in-the-loop validation should be performed. This can be conducted by regularly involving human experts to review samples of system outputs to ensure ongoing alignment with expert judgment.

Finally, it is important to ensure bias is not occurring in the system. This can be achieved by implementing specific checks to ensure the system maintains fairness across different demographic groups and content types.

By incorporating these measures, particularly human-in-the-loop and validation and bias prevention, the REGAI system can maintain its performance and fairness over time. This addresses the crucial need for long-term reliability of AI implementations.

## 5.2. Explainability Mechanism in REGAI

Explainability is crucial for the widespread adoption and trust in the REGAI system. This is important for most of the applications discussed in Section 4, given their high-stakes nature, such as academic grading and job candidate evaluation. The explainability mechanism serves several key purposes.

First, it provides transparency by providing insight into how the system arrives at its evaluations. This allows users to understand and evaluate the reasoning behind each score.

Second, it promotes trust by making the decision-making process transparent. It helps build trust among users, whether they are educators, hiring managers, or other professionals relying on the system's outputs.

Third, it facilitates error detection. Explainable outputs make it easier to identify when the system is making a mistake or deviating from the intended rubric application.

Finally, it drives continuous improvement. Clear explanations facilitate feedback gathering from human experts, which can be used to refine the system and improve its alignment with human judgment.

The REGAI pipeline inherently incorporates several features that contribute to its explainability. These include using rubrics, providing justifications, and implementing a critique cycle. Detailed rubrics, which are human-created or human-reviewed, provide a clear framework for evaluation that can be referenced in explanations. Justification must be provided by the scoring agent in the form of a detailed explanation for each score, which explicitly references the rubric criteria. The critique cycle facilitates a back-and-forth interaction between the scoring and critiquing agents, creating a visible trail of reasoning and refinement.

To further enhance explainability, several additional mechanisms could also be implemented. These include language highlighting, critique summarization, comparison visualization, using confidence metrics, performing continuous assessment and leaving an audit trail.

Language highlighting is a system that identifies specific language in the evaluated text that corresponded to the rubric criteria. This visually links the text to the assessment.

A critique summary highlights the key points raised during the critique cycle, showing how the initial assessment was refined. Comparison visualization can be leveraged when using retrieval-augmented generation (RAG) to provide visualizations of how the current evaluation compares to similar cases from the knowledge base. Confidence metrics, which are measures of the system's confidence in its assessment for each criterion, can be used to flag areas where human review might be particularly valuable. Continuous assessment and review can be performed by having expert judges review samples from the model on a continuous basis, or when anomalies are detected.

Finally, an audit trail is a comprehensive log of the steps in the evaluation process, from initial scoring through each iteration of the critique cycle. It can be accessed for in-depth review if needed.

It's important to note that while striving for maximum explainability is laudable, care must be taken to balance this with system requirements such as processing speed and output conciseness. The level of detail in explanations is adjustable based on the specific use case and user needs.

## 6. Data and Analysis

This section presents a preliminary evaluation of the REGAI system, comparing its performance across two configurations: a REGAI critique cycle with a single critique iteration and a rubric-only approach. These initial results provide a proof of concept and are indicative, rather than definitive, of the system's potential capabilities.

### 6.1 Methodology

Evaluation was conducted using a dataset of the first 500 essays from essay set eight of the Hewlett Foundation Automated Essay Scoring Dataset [48]. Each essay was scored by the REGAI system under two conditions:

1. A single critique iteration configuration.
2. A rubric-only approach without the critique cycle.

The first method is a preliminary implementation of REGAI with a limited number of human-approved examples. For the second, the LLM is simply given the rubric and asked to score the essay based on it.

The automated scores were then compared against human-provided scores, which served as the basis for this evaluation. The analysis focuses on three key aspects: overall performance metrics, category-wise performance, and score distribution.

Each essay in the Hewlett Foundation essay set eight evaluation data contains six category scores: ideas and content (I&C), organization (Org), voice, word choice (WC), sentence fluency (SF), and conventions (Conv). Each has a score range from 1-6. However, only I&C, Org, SF, and Conv factor into the overall score. Conv has twice the weight of the other three categories in the score, resulting in a total score out of 30. Then, a final score – consisting of the two scorers summed scores – is given, which is out of a total of 60. For this analysis, REGAI was asked to score each of the categories, and calculate the final score. This calculation likely distorted the mean average error metric, as any errors between the initial scorer(s) and the REGAI system grader would be amplified by 2.

For this testing, only two examples of grades and two examples of critiques of grades were available as examples in the system. As such, each grade and critique got the same two examples. As will be seen in the correlation analysis, this is likely to have caused some overfitting. The development of the knowledge base will be a priority for future versions of REGAI.

**6.2 Overall Performance Metrics**

The overall performance of both configurations was assessed using three key metrics: correlation with human scores (Corr), mean absolute error (MAE), and quadratic-weighted kappa (QWK). These metrics characterize the system's accuracy and reliability. These metrics are provided in Table 1. SC refers to the single critique method, while RO denotes the rubric-only approach.

Table 1. REGAI Performance Metrics.

| Metric | SC | RO | Raters | Diff (SC - RO) | Diff (SC - Raters) | Diff (RO - Raters) |
|---|---|---|---|---|---|---|
| Correlation | 0.49 | 0.55 | 0.60 | -0.06 | -0.10 | -0.04 |
| MAE | 4.37 | 7.86 | 2.03 | -3.49 | 2.34 | 5.83 |
| QWK | 0.49 | 0.26 | 0.59 | 0.23 | -0.10 | -0.33 |

Table 1 also presents a comparison of the metrics between the human raters. The results reveal several interesting findings.

First, in terms of correlation, the rubric-only approach showed a slightly higher correlation (0.551) with human scores, as compared to the single critique method (0.495). However, both methods fall short of the correlation between human raters (0.595). This result suggests that, while both automated methods show promise, there is still room for improvement in aligning with human scoring patterns.

For the mean absolute error metric, the single critique method demonstrated a substantially lower MAE (4.37) as compared to the rubric-only approach (7.86). Interestingly, both automated methods have a higher MAE than the human raters (2.032), indicating that, while the SC method improves upon RO, there is still a gap in accuracy compared to human raters.

The single critique method achieved a higher QWK (0.489), as compared to the rubric-only approach (0.260). However, both methods fall short of the inter-rater QWK (0.588). This suggests that, while the critique cycle enhances the system's ability to distinguish between different levels of essay quality, there is still room for improvement in matching human judgment consistency.

These initial results present a mixed picture. The single critique method shows clear advantages over the rubric-only approach in error reduction and quality differentiation. However, both methods still have room for improvement to reach human raters' performance.

### 6.3 Category-wise Performance

Next, analysis of behavior across the six categories scored in the essay evaluation dataset was performed. These areas are ideas and content (I&C), organization (Org), voice, word choice (WC), sentence fluency (SF), and conventions (Conv). Table 2 presents the category-wise correlations (Corr), mean absolute errors (MAE), and quadratic weighted kappa (QWK) for both automated grading configurations and human raters.

Table 2. REGAI Category-Wise Performance Metrics.

| Category | SC Corr | RO Corr | SC MAE | RO MAE | SC QWK | RO QWK | Rater Corr | Rater MAE | Rater QWK |
|---|---|---|---|---|---|---|---|---|---|
| I&C | 0.42 | 0.43 | 0.90 | 0.46 | 0.27 | 0.41 | 0.50 | 0.45 | 0.50 |
| Org | 0.30 | 0.44 | 0.60 | 0.70 | 0.29 | 0.31 | 0.49 | 0.43 | 0.49 |
| Voice | 0.34 | 0.34 | 0.56 | 0.52 | 0.31 | 0.33 | 0.41 | 0.44 | 0.40 |
| WC | 0.37 | 0.37 | 0.53 | 0.91 | 0.33 | 0.18 | 0.47 | 0.37 | 0.46 |
| SF | 0.34 | 0.36 | 0.64 | 1.06 | 0.28 | 0.16 | 0.48 | 0.43 | 0.48 |
| Conv | 0.30 | 0.31 | 0.83 | 1.43 | 0.17 | 0.09 | 0.51 | 0.43 | 0.51 |

In terms of correlation patterns: across most categories, the rubric-only approach showed marginally higher correlations with human scores. However, both automated methods generally show lower correlations than the inter-rater correlations, indicating room for improvement in capturing the nuances of each category.

For MAE, the single critique method demonstrated a lower measure for most categories, as compared to the rubric-only approach. However, human raters consistently produced lower MAE across all categories. This suggests that, while the critique cycle improves accuracy, there's still a gap in matching human-level precision.

Finally, for QWK, the SC method generally showed higher performance than the RO method, but both fall short of the inter-rater QWK, in most categories. This suggests that, while the critique cycle improves agreement with human raters, there's still room for enhancing consistency across categories.

These category-wise results highlight the complexity of automated essay scoring and the need for a nuanced approach in implementing and refining the critique cycle across different evaluation criteria.

### 6.4. Score Distribution Analysis

To further understand how the REGAI system compares with human scoring patterns, the distribution of scores across both configurations and human-resolved scores (HRS) were analyzed. Table 3 provides descriptive statistics for these score distributions.

Table 3. REGAI Score Distribution Statistics.

| Metric | SC | RO | HRS |
| --- | --- | --- | --- |
| Count | 500 | 500 | 500 |
| Mean | 36.17 | 29.30 | 36.95 |
| Std | 5.56 | 4.99 | 5.56 |
| Min | 17 | 15 | 15 |
| 25% | 33 | 26.92 | 33 |
| 50% | 37 | 30 | 37 |
| 75% | 40 | 32.25 | 40 |
| Max | 50 | 42 | 60 |

The mean score for the single critique method (36.17) closely aligns with the human-resolved mean (36.95), while the rubric-only approach had a lower mean (29.30). This suggests that the critique cycle helps calibrate the overall scoring level to better match human raters.

Both REGAI configurations have a narrower score range, as compared to human scores. The single critique method ranges from 17 to 50, while human scores span from 15 to 60. This compression of the score range is a common challenge in automated scoring systems. This may indicate a need for further calibration.

The standard deviations of the single critique method (5.56) and human-resolved scores (5.56) are nearly identical. This indicates a similar spread of scores. The rubric-only approach shows a slightly lower standard deviation (4.99), suggesting a tendency to cluster scores more tightly around the mean.

These distribution patterns indicate that the single critique method achieves a closer approximation of human scoring patterns, in terms of central tendency and score spread. However, there is still room for improvement in capturing the full range of scores, as compared to the human graders.

## 6.5. Statistical Significance

To assess the statistical significance of the difference between the two REGAI configurations, a t-test was conducted on their respective overall scores. Table 4 presents the results of t-tests comparing the different scoring methods.

**Table 4.** Statistical Significance of Difference Between REGAI Method Evaluations.

| Comparison | t-statistic | p-value |
| --- | --- | --- |
| SC vs RO | 20.57 | $1.24 \times 10^{-78}$ |
| SC vs R1 | 62.79 | 0 |
| RO vs R1 | 41.82 | $1.30 \times 10^{-221}$ |
| R1 vs R2 | -1.194 | 0.23 |

The extremely low p-values for comparisons involving the automated methods (SC and RO) indicate that the differences in scores between these methods and human raters are statistically significant. This suggests that, while automated methods show promise, they are still distinguishable from human scoring patterns. Interestingly, the comparison between the raters shows no statistically significant difference (p-value > 0.05), demonstrating the consistency between the human raters.

## 6.6. Discussion and Limitations

The results presented offer insights into the potential of the REGAI system. In particular, they demonstrate the benefit of incorporating a critique cycle. The single critique method has the clear advantage of reducing error magnitude and improving the differentiation of essay quality, as evidenced by the lower MAE and higher QWK values. However, the lower correlation with human scores, as compared to the rubric-only approach, suggests that there is room for refinement of how the critique cycle aligns the final scores with human judgment.

Several limitations that impacted this experimentation should be noted. First, the evaluation only considered a single critique iteration. Future work can explore the impact of multiple critique cycles on performance. Second, the evaluation was conducted on a single set of 500 essays. The system's performance may vary with different essay types and subject matters. Finally, there is substantial room for enhancement of various aspects of the system. These include prompt engineering, knowledge base expansion, and fine-tuning of the critique process.

Despite these limitations, the results presented herein provide encouraging evidence for the potential of the REGAI approach. In particular, they show the value of incorporating a critique cycle in automated essay scoring. The system's ability to reduce error magnitude and improve quality differentiation suggests that, with further refinement, it could offer a powerful tool for large-scale, consistent, and nuanced text evaluation.

# 7. Conclusion and Future Work

The rubric enabled generative artificial intelligence system presented in this paper is an approach to the automated evaluation of unstructured text data. By combining the power of large language models, retrieval-augmented generation, and human expertise, REGAI addresses many of the challenges associated with current AI-based evaluation systems. In particular, it provides benefits in terms of alignment with human judgment, explainability, and adaptability across diverse domains.

The key strengths of REGAI lie in its iterative critique cycle, which promotes adherence to human-defined rubrics, and its self-reinforcing knowledge base, which allows the system to continuously improve its performance over time. The system's versatility has been demonstrated through its potential applications in various fields, including academic grading, human resources, professional training, and legal document analysis.

The development, implementation and evaluation of REGAI have also highlighted several areas that require further research and development.  Comprehensive empirical studies are needed to quantitatively assess its performance across different domains and compare it with existing evaluation methods and human experts. Additionally, further work is required to address potential biases in the system and ensure fairness in high-stakes decision-making processes, particularly for applications like job candidate evaluation and legal analysis.

Continuous monitoring of the pipelines' results should be implemented to find biases, if they exist. Research should also be performed to optimize the critique cycle and knowledge base. This will be crucial for deploying REGAI in large-scale, real-time applications.

Investigating techniques to efficiently adapt REGAI to new domains with minimal expert input could significantly broaden its applicability. Exploring the integration of REGAI with other AI technologies, such as computer vision and speech recognition, could extend its capabilities to multimodal evaluation tasks (e.g., videos of client interactions could be included with the transcript for evaluation).

Developing robust methods for tracking and maintaining REGAI's performance over extended periods, especially as language models and evaluation standards evolve, will be essential for its long-term viability. Finally, creating intuitive interfaces for experts to interact with REGAI, particularly in rubric creation and review processes, will be crucial for its adoption in various professional settings.

While REGAI is a promising framework for automated text evaluation, its full potential can only be realized through additional research, rigorous testing, and thoughtful implementation. As AI continues to evolve, systems like REGAI have the potential to significantly enhance decision-making processes across numerous fields, provided they are developed and deployed with careful consideration of their limitations and ethical implications.

# Appendix A. System Overview.

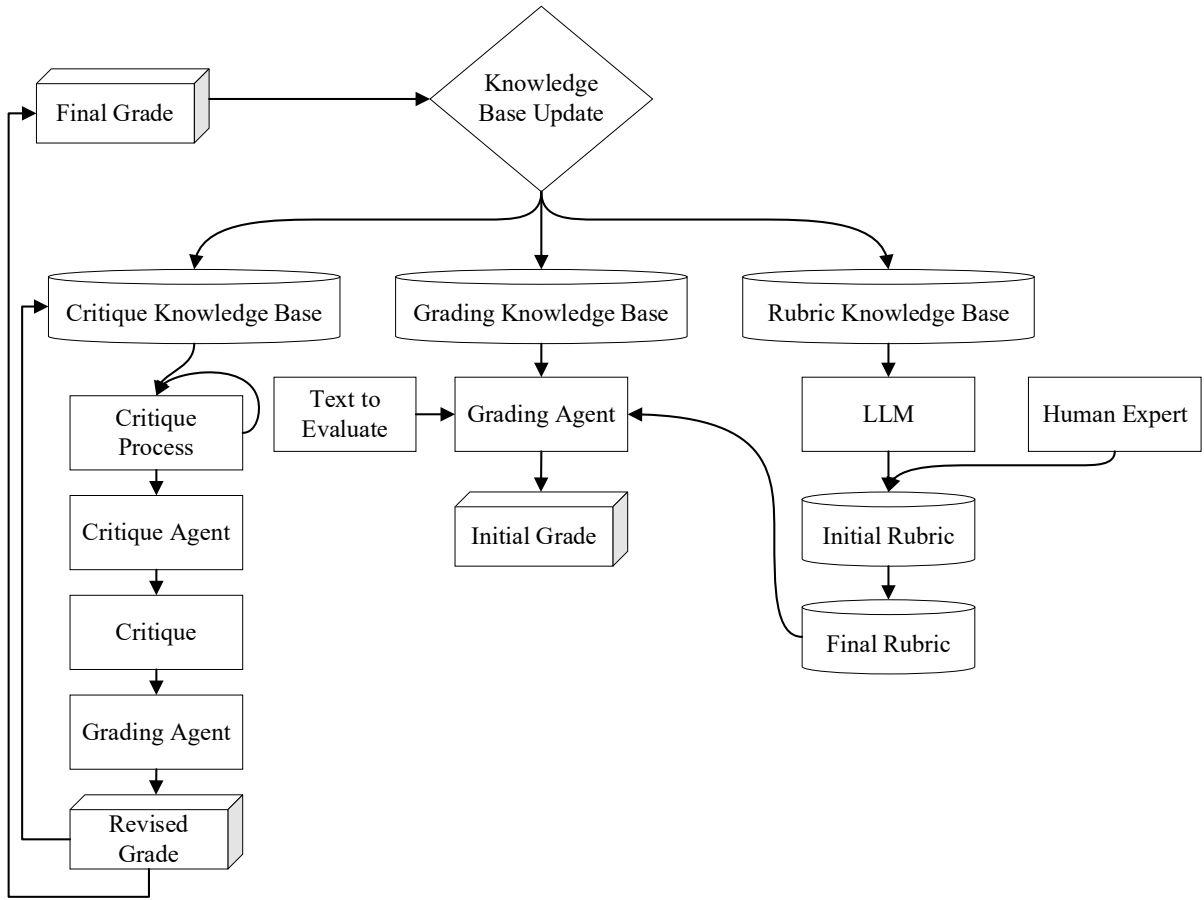

**Figure A1.** Comprehensive REGAI System Overview.